\documentclass{article}
\usepackage[utf8]{inputenc}
\usepackage{graphicx}
\usepackage{natbib}
\usepackage{hyperref}
\usepackage{amsmath}
\usepackage{amssymb}
\usepackage[capitalize,noabbrev]{cleveref}
\usepackage[hide=false,marginparwidth=1in]{marginalia}
\usepackage{comment}
\usepackage{fullpage}
\usepackage{authblk}

\begin{document}

\title{\textbf{Dataset Difficulty and the Role of Inductive Bias}}
\author[1,2]{Devin Kwok}
\author[1,2,3]{Nikhil Anand}
\author[4]{Jonathan Frankle}
\author[5,1,2]{Gintare Karolina Dziugaite}
\author[1,2]{David Rolnick}
\affil[1]{McGill University}
\affil[2]{Mila - Quebec AI Institute}
\affil[3]{Amazon, Alexa AI\footnote{Work done prior to joining Amazon.}}
\affil[4]{MosaicML\footnote{Work done prior to joining MosaicML.}}
\affil[5]{Google DeepMind}

\maketitle

\begin{abstract}
    Motivated by the goals of dataset pruning and defect identification, a growing body of methods have been developed to score individual examples within a dataset.
    These methods, which we call ``example difficulty scores'', are typically used to rank or categorize examples, but the consistency of rankings between different training runs, scoring methods, and model architectures is generally unknown.
    To determine how example rankings vary due to these random and controlled effects, we systematically compare different formulations of scores over a range of runs and model architectures.
    We find that scores largely share the following traits: they are noisy over individual runs of a model, strongly correlated with a single notion of difficulty, and reveal examples that range from being highly sensitive to insensitive to the inductive biases of certain model architectures. Drawing from statistical genetics, we develop a simple method for fingerprinting model architectures using a few sensitive examples.
    These findings guide practitioners in maximizing the consistency of their scores (e.g. by choosing appropriate scoring methods, number of runs, and subsets of examples), and establishes comprehensive baselines for evaluating scores in the future.
\end{abstract}

\section{Introduction}

As artificial neural networks and the datasets used to train them have rapidly increased in size and complexity, it has become increasingly important to identify which examples affect learning and how. 
Indeed, in recent years a growing number of proposals have appeared for scoring and ranking individual examples in the training set in order to identify which examples may suffer from catastrophic forgetting  \citep{toneva2018an}, are affected by network pruning \citep{jin2022pruning}, contribute to improved generalization \citep{paul2021deep, nohyun2023data, sorscher2022beyond, swayamdipta2020dataset}, improve downstream model robustness \citep{northcutt2021confident, jiang2020characterizing}, or help in mitigating biases in the model \citep{hooker2019compressed, hooker2020characterising}.
A common theme among these scores is that the most ``difficult'' (e.g. highest error) examples often significantly affect learning, while a substantial fraction of ``easy'' (lowest error) examples can be discarded without adversely affecting model properties such as test accuracy. 
Beyond direct applications such as data pruning, where a fraction of a large dataset can be used to train models to equivalent performance \citep{toneva2018an, hooker2020characterising, paul2021deep, nohyun2023data, sorscher2022beyond}, these methods can be used to identify outliers \citep{carlini2019distribution, feldman2020neural, siddiqui2022metadata, paul2021deep}, find underrepresented classes of examples \citep{hooker2020characterising}, correct for biases and spurious correlations \citep{yang2023identifying}, remove mislabelled or redundant examples \citep{pleiss2020identifying, birodkar2019semantic, Pruthi2020, jiang2020characterizing, swayamdipta2020dataset}, and prioritize important examples in training through curriculum-based learning \citep{paul2021deep, jiang2019accelerating}.

There exists a plethora of scores targeting various applications, but regardless of their uses, one needs a good grasp of their properties and robustness in order to use scores appropriately.
We thus take the application of each score out of the question, and instead evaluate how scores vary between runs, how they relate to one another, and how transferable they are between different model architectures.
Since each score implicitly or explicitly ranks training examples by their ``difficulty'', we refer to these scores as \emph{example difficulty scores}.
That said, the difficulty of an example cannot be defined independently, since scores vary depending on the rest of the training data, the inductive biases of the model, and the training algorithm.
Many scores also require additional information from the data or model such as gradient information \citep{paul2021deep}, gradient variance among the training data \citep{agarwal2020estimating}, the training trajectory \citep{siddiqui2022metadata}, or the error signal \citep{toneva2018an, swayamdipta2020dataset}.

From a research standpoint, our study sets a foundation for comparing example difficulty scores by establishing a proper evaluation protocol and baselines against which new scores can be compared.
In particular, we ask \textit{how much do example rankings vary due to random effects} (e.g. model initialization), and \textit{do different scores measure independent notions of difficulty} (i.e. how correlated are their example rankings)?
From a practical standpoint, our answers to these questions may help practitioners determine which scores to choose based on the downstream use and their desired robustness/invariance (e.g., to architectures), how many runs are needed to rank examples reliably, and whether a compute-costly score can be interchanged for a cheaper one.
A practitioner may also want to know if rankings are consistently affected by changes to the model (e.g., the size of a neural network) or hyperparameters in the learning algorithm (e.g., learning rate in stochastic gradient methods) \cite{paul2021deep,jiang2020characterizing}---properties which we call the \textit{inductive biases} of the model.
To address this question, we consider how example difficulty changes in response to different inductive biases, and whether this difference is detectable above the random variation due to different runs.
We summarize our findings below.

\begin{figure}
    \centering
    \vspace{-0.15in}
    \includegraphics[width=0.3\textwidth]{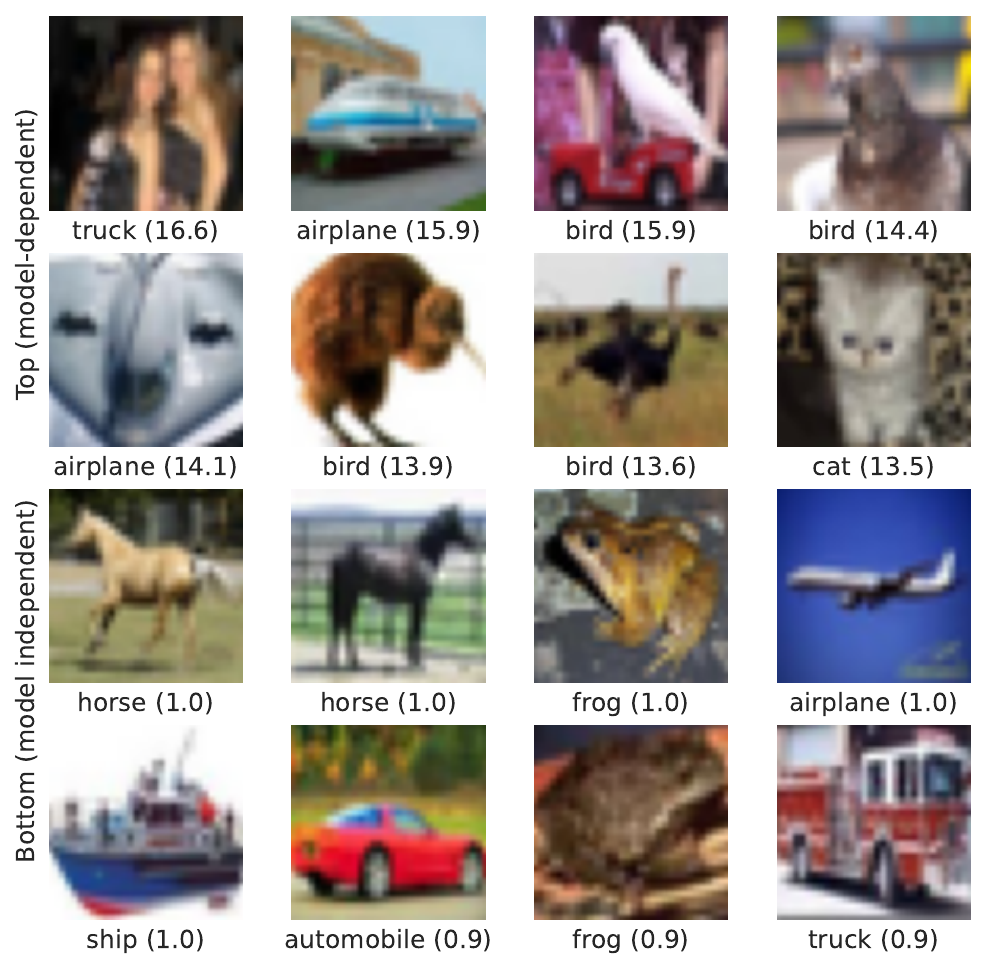}
    \includegraphics[width=0.5\textwidth]{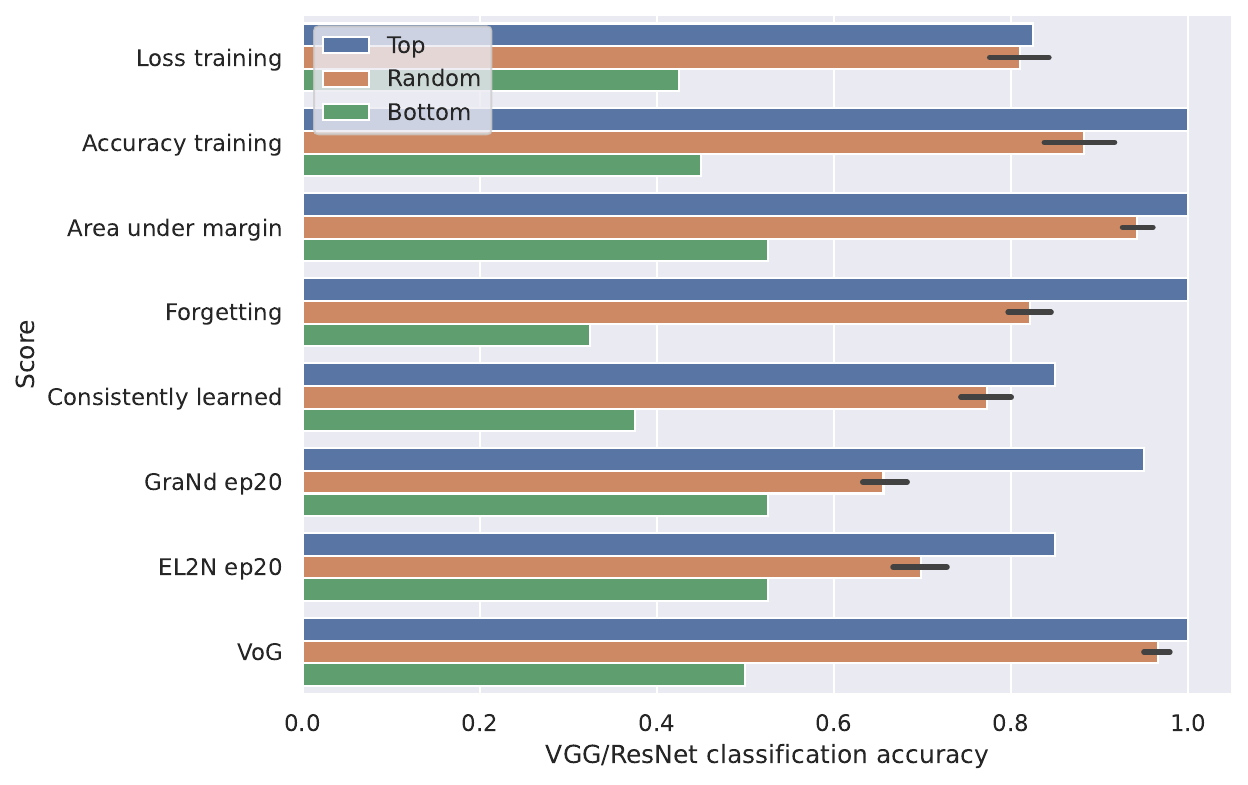}
    \caption{
    \textbf{Left:} Top 8 CIFAR-10 examples most/least sensitive to inductive biases (model width, depth, and architecture), ranked by statistical significance of changes to difficulty score for pairwise model comparisons (see \cref{sec:bias} for more details). Numbers indicate mean $-\log P$ value of each example.
    \textbf{Right:} Distinguishing between VGG and ResNet using the top 8 examples (Left) as features for logistic regression.
    }
    \label{fig:inductive-bias-examples}
\end{figure}

\paragraph{Key contributions:}
\begin{enumerate}
\item
    \label{keycontribution1}
    There is substantial variance in difficulty between different runs (due to random initialization, etc.), and this variation is generally higher for harder examples (\cref{fig:variance}).
    As a result, \textbf{ranking or thresholding examples is unreliable when scores are averaged over a small number of runs} (\cref{fig:rank-threshold}).
\item
    \textbf{All difficulty scores are generally well correlated} with themselves and other scores (\cref{fig:covariance}). Averaging over multiple runs almost always improves this correlation (gradient-based scores are an exception).
\item
    \textbf{A single principal component captures most of the variation in difficulty} among scores computed on training examples (\cref{fig:pca-components}).
    The remaining variation is independent between scores and cannot be reduced by linear combination to make data ranking/thresholding more consistent.
\item
    Different types of models with distinct inductive biases cause \textbf{statistically significant changes in difficulty for many examples}.
    Scores from as few as 8 examples can be used as a fingerprint to reliably distinguish between inductive biases (\cref{fig:inductive-bias-examples}).
\end{enumerate}

These results highlight several underappreciated aspects of using difficulty scores in practice.
First, ranking or classifying individual examples (e.g. to identify mislabelled examples) is error-prone, since the per-run variance of scores is high and differs from example to example.
The cost of training large models may additionally limit the number of runs (e.g. 5-10) on which scores are averaged, worsening this issue.

Second, all scores are fairly predictive of one another in spite of their methodological differences.
A key motivation for many scores is to be a proxy for other, more computationally-intensive scores \cite{paul2021deep,jiang2020characterizing,pleiss2020identifying,agarwal2020estimating,baldock2021deep,sorscher2022beyond}.
Our work shows that the correlation between proxy scores needs to be quite high to beat the baseline similarity among unrelated scores.

Finally, the role of the model has typically been ignored or averaged out in practice.
Drawing inspiration from statistical genetics, we show that models affect example difficulty in predictable ways.
We find that a handful of examples with the most significant changes can reliably distinguish between inductive biases, giving practitioners a new tool for fingerprinting different models.
Conversely, practitioners should be wary of comparing scores from different models, since only a minority of examples have model-independent scores.

\section{Methodology \label{sec:overview}}

We seek to tease apart the influence of various factors on example difficulty: namely, the role of uncontrolled/random effects (e.g. model initialization), the scoring method, and controlled effects (e.g. model architecture). These factors naturally divide our experiments into measuring the variance, covariance, and bias of scores.

\subsection{Variance}

Scores are often averaged over many runs to increase their consistency.
However, with large models the number of runs is often limited by computational resources.
In this situation a practitioner might ask: how many runs correspond to a given level of error in estimated score, and given a choice of scores, which ones are least affected by per-run variance?

\begin{figure}
    \centering
    \includegraphics[width=\textwidth]{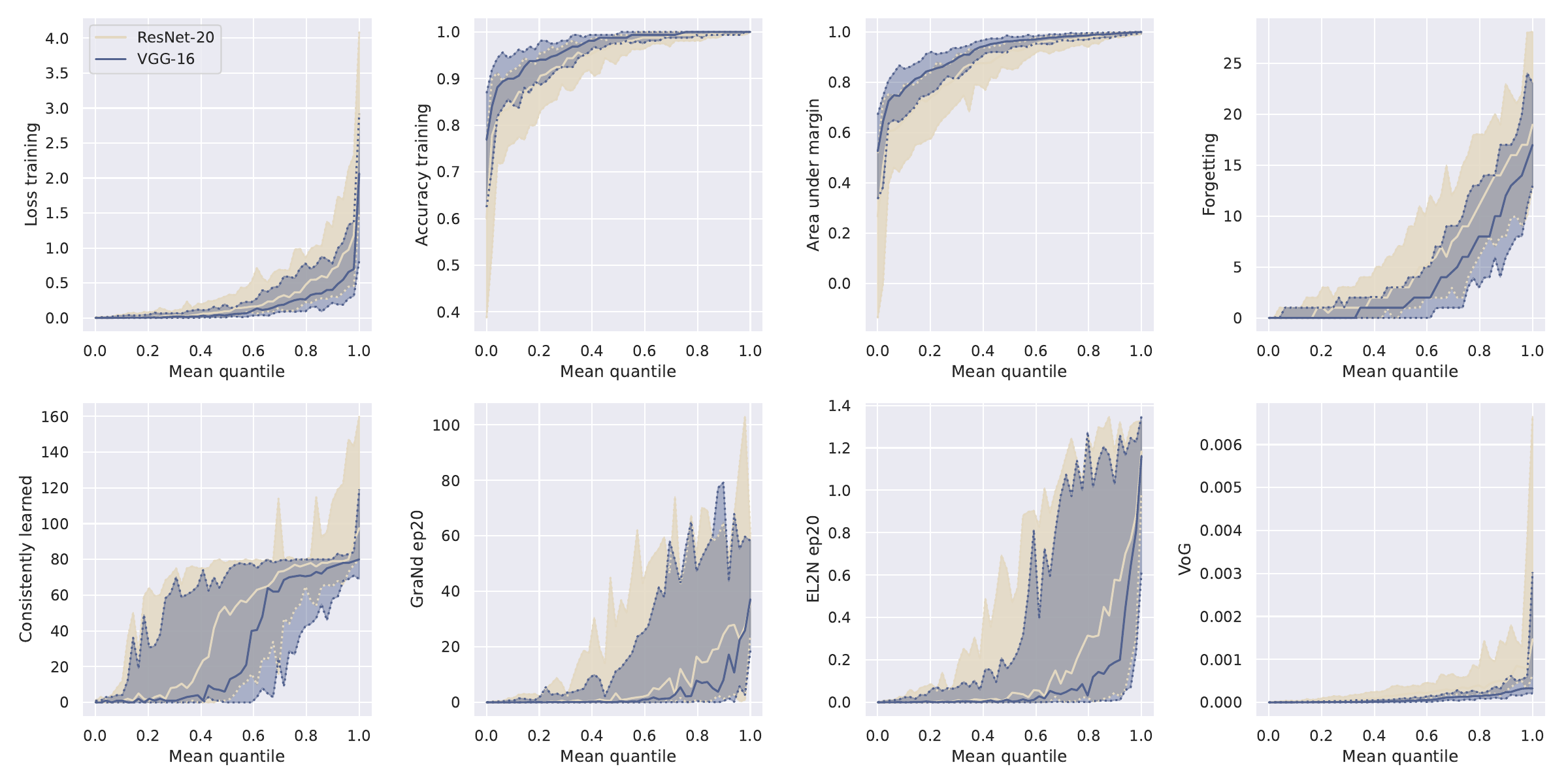}
    \\
    \caption{
    Variance of selected difficulty scores for CIFAR-10 examples.
    \textbf{Rows 1-2:} Score distribution for individual examples (y-axis) and individual runs, ordered by the rank of their mean score (x-axis). To make the plots more legible, examples are binned into groups of 1000. Shaded region indicates 90\% confidence interval. From top to bottom and left to right: mean loss over training, mean accuracy over training, area under margin (correct class probability minus top other class) over training, number of forgetting events, epoch when example is learned without further forgetting, gradient norm at epoch 20, $L^2$ norm of output probability error at epoch 20, mean per-pixel variance of gradients at input over training.
    }
    \label{fig:variance}
\end{figure}

While prior work shows that some scores are consistent over different runs \citep{baldock2021deep}, such as by calculating between-run correlations \citep{toneva2018an}, a baseline is needed to understand what degree of consistency is to be generally expected.
Furthermore, the degree to which some examples are less consistent than others is unknown. To address these issues, we plot the distribution of per-run score variance for all examples (\cref{sec:var}, \cref{fig:variance}) to understand how score error changes at different levels of difficulty, and the effects of this error in the two common applications of splitting and ranking data (e.g. for data pruning, outlier detection, and training prioritization).

For extremely large models where multiple runs are impractical (e.g. \citet{dosovitskiy2021an}), practitioners may also want to know whether individual runs can reliably estimate the average score of an example, or predict other types of scores or the outcome of other runs.
To address this, we compute pairwise correlations between scores on a mean-to-mean, mean-to-run, and run-to-run basis (\cref{sec:cov}, \cref{fig:covariance}) to cover all combinations of predictions between scores from single/multiple runs.

\subsection{Covariance}

The large variety of difficulty scores with distinct formulations makes it difficult to understand which scores are appropriate for a given task.
Simple scores are also preferred over compute-expensive scores (such as those averaged over an ensemble of models) if they are more or less equivalent.
Practitioners may therefore ask: how do different scores relate to each other, and do they measure the same notion(s) of difficulty?

Previous works address these questions to a limited extent, such as the error norm (EL2N) being strongly correlated with the gradient norm (GraNd) and forgetting score \citep{paul2021deep}, or prediction depth correlating well with example learning times and adversarial margins \citep{baldock2021deep}.
Again however, a baseline is needed to understand what score substitutions are possible and the general level of similarity expected between scores.
The pairwise correlations between scores in \cref{fig:covariance} establish this baseline, enabling existing and newly proposed scores to be compared by their similarity and ability to predict other scores.

To better understand what scores are actually measuring, practitioners may also want to know if there are multiple ways that examples can vary in difficulty, such as by counting the orthogonal directions covered by different scores.
Prior works have tended to propose two dimensions of difficulty \citep{hooker2019compressed, carlini2019distribution, baldock2021deep, swayamdipta2020dataset}, often related to train versus test set performance, but it is unknown if the dimensions proposed by these works are equivalent.
Although we only consider training examples in this work, we make a first step towards quantifying the dimensionality of difficulty scores by measuring the principal components of variation spanned by a variety of scores, and the degree to which different scores share principal components (\cref{sec:cov}, \cref{fig:pca-components}).

\subsection{Bias}

The difficulty of an example depends on both properties of the example \textit{and} the function class used to model it, such as the architecture of the model. For example, it is clear that a simple logistic regression and a large convolutional network would produce very different difficulty scores, and that their relative performance could shift drastically on images versus linearly separable data. However, if two models have comparable average performance, the degree to which difficulty scores evaluated on each model differ is not so obvious.

Prior work highlights how some scores tend to \textit{not} vary among different model architectures \citep{carlini2019distribution, paul2021deep, baldock2021deep}, but a comprehensive evaluation of the robustness of scores to different model architectures is lacking.
We also note that model-dependent scores may be desirable for some applications, such as identifying the inductive biases of a model.
To better understand the full range of model-dependent effects, we ask: \textit{which} scores and examples are more sensitive to a model's inductive bias, and conversely, which ones are more model-agnostic?
We address these questions by conducting a large-scale statistical analysis of the change in score between different model architectures at the level of \textit{individual} examples (\cref{sec:bias}, \cref{fig:bias-pvalue}).
This allows us to find particular examples which are extremely sensitive/insensitive to inductive biases, which can be used as a ``fingerprint'' to classify model architectures with only a few scores (\cref{fig:inductive-bias-examples}).

\section{Difficulty Scores and Related Work}

We provide a brief overview of difficulty scores, focusing on their conceptual relations.
Since the variance of a score depends strongly on sample size, we categorize scores by their sampling method:

\begin{itemize}

\item \textbf{Scores computed at a single point in training.}
    This category includes typical performance metrics computed at the end of training, such as accuracy.
    Since large models effectively have zero training loss, more sophisticated scores are needed to differentiate examples, including prediction depth \citep{baldock2021deep}, adversarial robustness \citep{carlini2019distribution}, and normalized margin from decision boundary \citep{jiang2018predicting}.
    Scores can also be computed during training for applications such as data pruning, including
    the $L^2$ norm of the gradient with respect to model parameters, or the norm of the output error \citep{paul2021deep}.
    Although scores in this category can also be averaged over training, they are typically evaluated at specific times for a specific purpose such as model evaluation.

\item \textbf{Scores computed throughout training.}
    These are scores requiring multiple model evaluations throughout training.
    Sampling methods include evaluating all examples at fixed intervals (e.g. per epoch), or evaluating each minibatch of examples when they are used in training \citep{toneva2018an}.
    Simple pointwise evaluations which can be averaged over all samples include loss \citep{jiang2020characterizing}, accuracy \citep{jiang2020characterizing}, entropy \citep{jiang2020characterizing}, model confidence \citep{jiang2020characterizing, carlini2019distribution, swayamdipta2020dataset}, and the classification probability margin \citep{toneva2018an,jiang2020characterizing,pleiss2020identifying}.
    Derived quantities include the variance of the above quantities over training \citep{swayamdipta2020dataset}, the number of forgetting events \citep{toneva2018an}, the time when an example is learned \citep{toneva2018an} or remains consistently learned \citep{baldock2021deep, siddiqui2022metadata}, and the use of entire loss trajectories as features for a downstream classifier \citep{siddiqui2022metadata}.
    Gradient-based methods include the variance of gradients (VoG) at the input \citep{hooker2020characterising}.

\item \textbf{Scores computed from an ensemble of models.}
    We omit scores from the previous categories since they can also be trivially averaged over an ensemble of models.
    Methods that average over different train/test splits include ensemble accuracy \citep{meding2021trivial}, train/test split consistency \citep{jiang2020characterizing}, influence and memorization \citep{feldman2020neural}, and retraining on held-out data \citep{carlini2019distribution}.
    Other scores include agreement in mutual information over an ensemble \citep{carlini2019distribution, kalimeris2019sgd}, robustness to $\epsilon$-differential privacy \citep{carlini2019distribution}, and true-positive agreement \citep{hacohen2020let}.
    Finally, scores that measure sensitivity to model perturbations include compression- and pruning-identified exemplars \citep{hooker2019compressed}.

\item \textbf{Training-free scores.}
    These scores are computed either with a closed-form expression or an arbitrarily chosen model, and are thus independent of the training process. Examples include the complexity gap \citep{nohyun2023data} and supervised/self-supervised prototypes \citep{sorscher2022beyond}.
\end{itemize}

As the list above is too broad to exhaustively analyze, we restrict the scope of our work as follows.
Firstly, we only look at models during training, excluding scores that require other model or data perturbations such as model pruning, model compression, or adversarial attacks.

Whether an example is seen by the model during training has obvious effects on its computed difficulty \citep{baldock2021deep, feldman2020neural, hacohen2020let, carlini2019distribution}. More subtly, the exact contents of the training dataset will also influence an example's difficulty \citep{feldman2020neural, carlini2019distribution}.
Our analysis uses default train/test splits and only scores training examples.
For ease of comparison, we include some precomputed scores from other works such as \citet{carlini2019distribution, feldman2020neural}.

Based on these restrictions, we select a representative subset of difficulty scores for the main text:
\begin{itemize}
    \item \textsc{Loss} and \textsc{Accuracy}: average loss and zero-one accuracy over training \citep{jiang2020characterizing}.
    \item \textsc{Area under margin}: the mean difference (over training) between the model's true class confidence (probability after softmax) and the highest confidence on any other class \citep{pleiss2020identifying}.
    \item \textsc{Forgetting} and \textsc{Consistently learned}: the number of times an example is forgotten (changes from correct to incorrectly classified) over successive minibatches \citep{toneva2018an}, and the first time an example is learned without further forgetting (also called iteration learned) \citep{baldock2021deep,siddiqui2022metadata}.
    \item \textsc{GraNd ep20} and \textsc{EL2N ep20}: the gradient norm and approximation of the norm as the $L^2$ norm of the error in output probability \citep{paul2021deep}. As these scores are used for early data pruning, we follow the original authors and compute them at epoch 20.
    \item \textsc{VoG}: (variance of gradients) measuring the mean variance of gradients with respect to individual input pixels \citep{agarwal2020estimating}.
    \item \textsc{Ensemble agreement} and \textsc{Holdout retraining}: also called ``agr'' and ``ret'' in \citet{carlini2019distribution}, these are precomputed scores for the Jensen-Shannon divergence between outputs in an ensemble, and the drop in KL-divergence when a trained model is fine-tuned on a hold-out dataset where the example in question is removed.
\end{itemize}


\section{Experiments and Results}

All figures in the main text are for scores computed with ResNet-20 \citep{he2016deep} models trained on CIFAR-10 \citep{Krizhevsky09CIFAR10}.
Difficulty scores are computed over 40 replicates for each model/dataset using standard SGD training for 160 epochs.

For experiments measuring inductive bias, we additionally run 20 replicates of variations on ResNet \citep{he2016deep} and VGG \citep{simonyan2015very} models with increased/reduced width/depth which are as follows: ResNet wide (ResNet-20-64, or $4 \times$ width), ResNet deep (ResNet-34), VGG narrow (VGG-16-16, or $1/4 \times$ width), and VGG shallow (VGG-11).

\subsection{Variance of Scores}
\label{sec:var}

We first examine the distribution of per-example scores over individual runs.
\cref{fig:variance} shows the 90\% confidence interval for scores binned by the quantile of their mean score over all runs.
A few general trends can be observed, namely that with a few exceptions (e.g. \textsc{Consistently learned}) easy examples both have less variation between runs and less variation between examples, whereas difficult examples tend to both have more variation and be more widely distributed.
This immediately indicates it is possible for all examples to vary in rank regardless of difficulty, because ranks are sensitive to the signal-to-noise ratio between each example's variance and the gap between the scores of nearby examples.
While some scores have less variance (i.e. narrower confidence intervals), scores with less variation also tend to be more tightly distributed (i.e. their plots in \cref{fig:variance} are flatter).
Thus, for all scores the majority of examples are inconsistently ranked.

\begin{figure}
    \centering
    \includegraphics[width=0.32\textwidth]{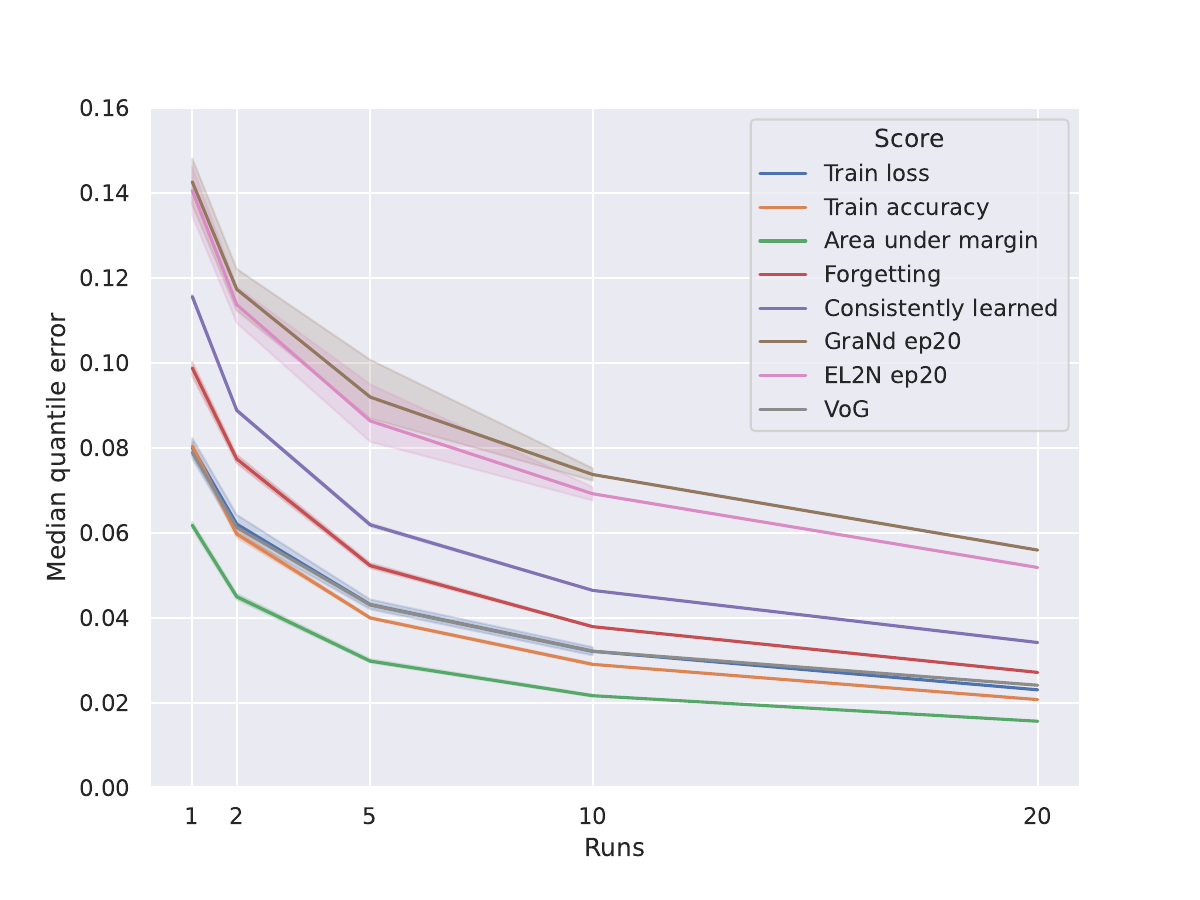}
    \includegraphics[width=0.32\textwidth]{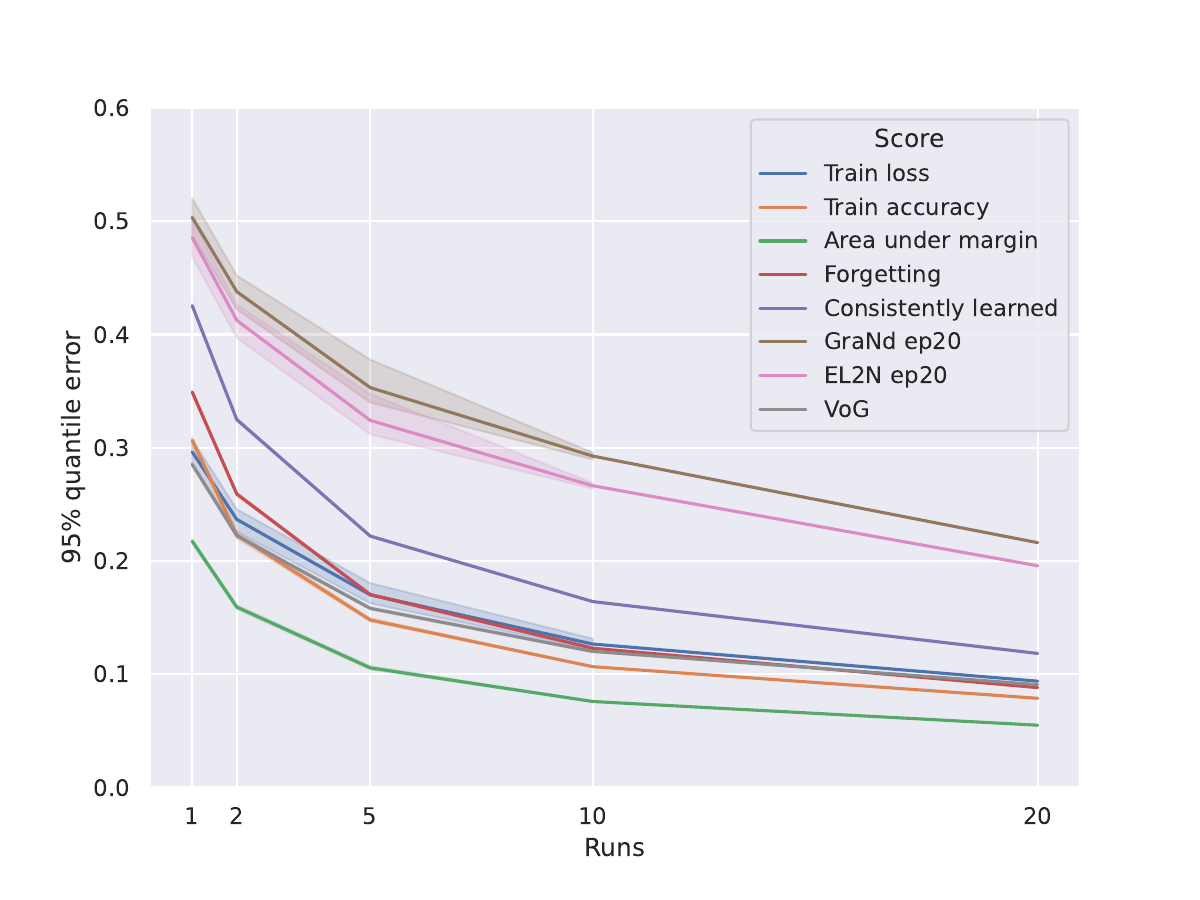}
    \includegraphics[width=0.32\textwidth]{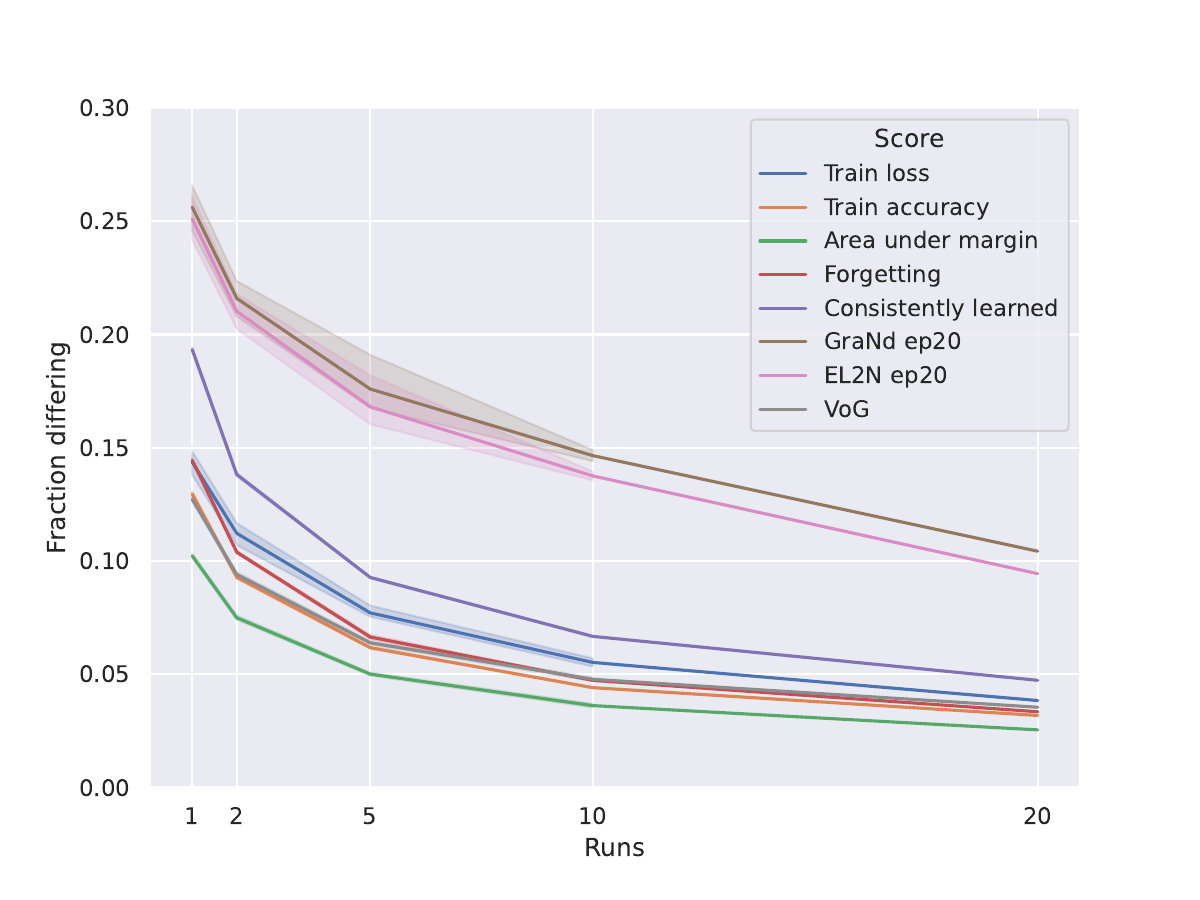}
    \caption{Variability in example difficulty scores due to sample size. \textbf{Left and Center:} Variability in ranking examples by their mean difficulty scores. Scores are averaged over a fixed number of runs (x-axis), and the median (Left) and 95\% confidence interval (Center) of absolute rank change between two such sets of runs is shown (y-axis).
    \textbf{Right:} variability in 50\% data splits found by thresholding mean difficulty scores. Number of runs that scores are averaged over (x-axis) is plotted against the fraction of examples that change classification between two such sets of runs (y-axis).
    }
    \label{fig:rank-threshold}
\end{figure}

To demonstrate the practical implications of this variance, we consider two applications: ranking and splitting datasets using scores averaged over some number of runs.
\cref{fig:rank-threshold} shows that the median difference in ranking between scores averaged over two sets of 5 runs can be 3-10\% of the dataset, with a 95\% confidence interval at 10-40\% of the dataset.
Similarly, using 5 runs gives 5-20\% disagreement between subsets of data split using a 50\% percentile threshold for scores.
These errors in turn introduce variation into the performance of downstream tasks such as data pruning or outlier detection using example ranks/thresholds.
Note that although multiple scores are presented together in \cref{fig:rank-threshold}, they are not directly comparable in variability since they have different methods and goals (e.g. \textsc{EL2N} and \textsc{GraNd} are only computed at a single epoch to allow for early data pruning).

\subsection{Correlation Between Scores}
\label{sec:cov}

We next examine the degree to which different scores are related, and whether scores averaged over runs are more or less predictive of individual run scores and other mean scores.
As scores generally have non-linear but monotonic relationships, we use Spearman (rank) correlation to better capture the full dependence between scores.
Note that correlations between related quantities are expected to be strong: for example, \textsc{Loss training}, \textsc{Accuracy training}, and \textsc{Area under margin} are all functions of the model output, although they are not directly recoverable from one another.

\begin{figure}
    \centering
    \includegraphics[width=0.32\textwidth]{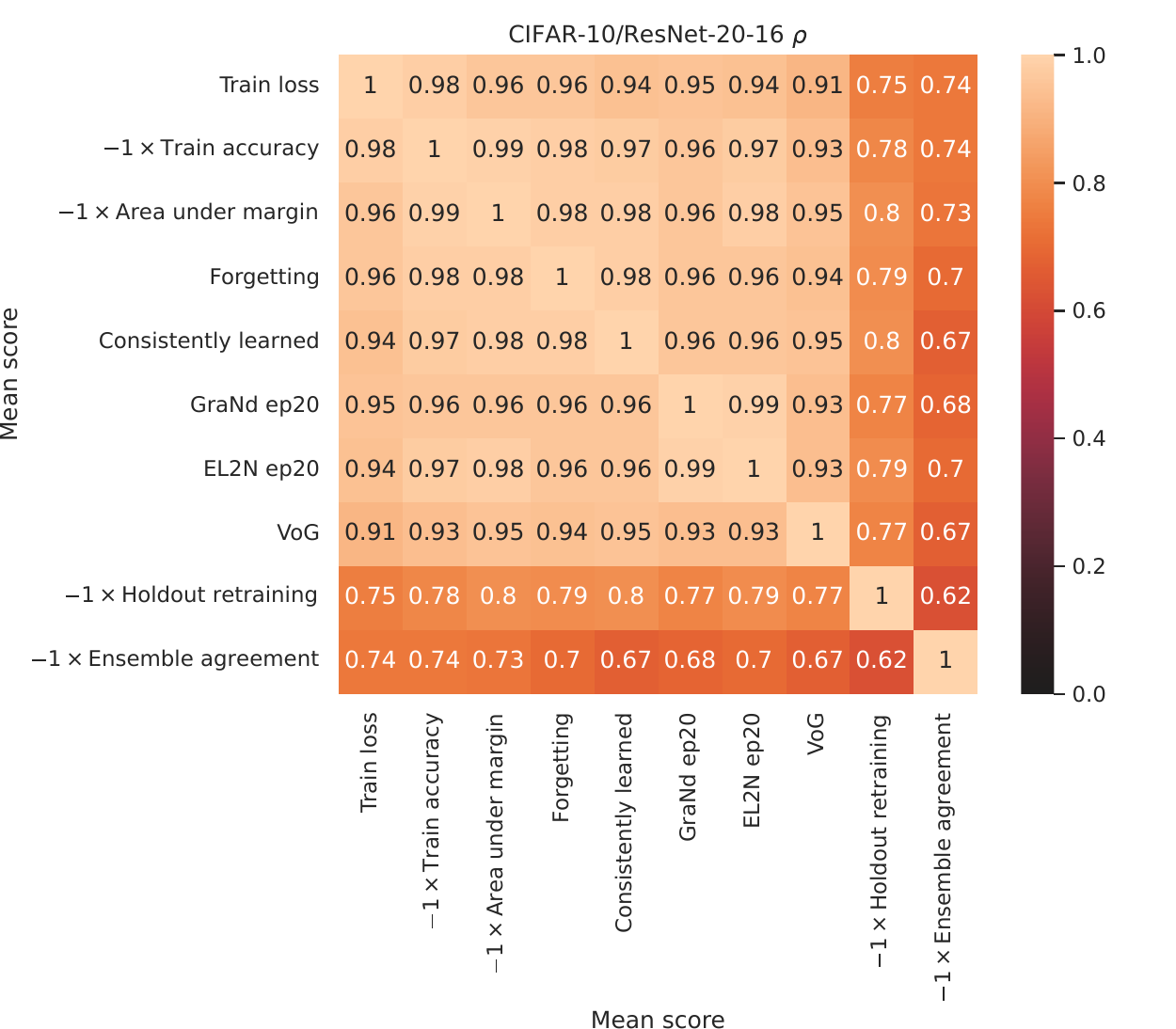}
    \includegraphics[width=0.32\textwidth]{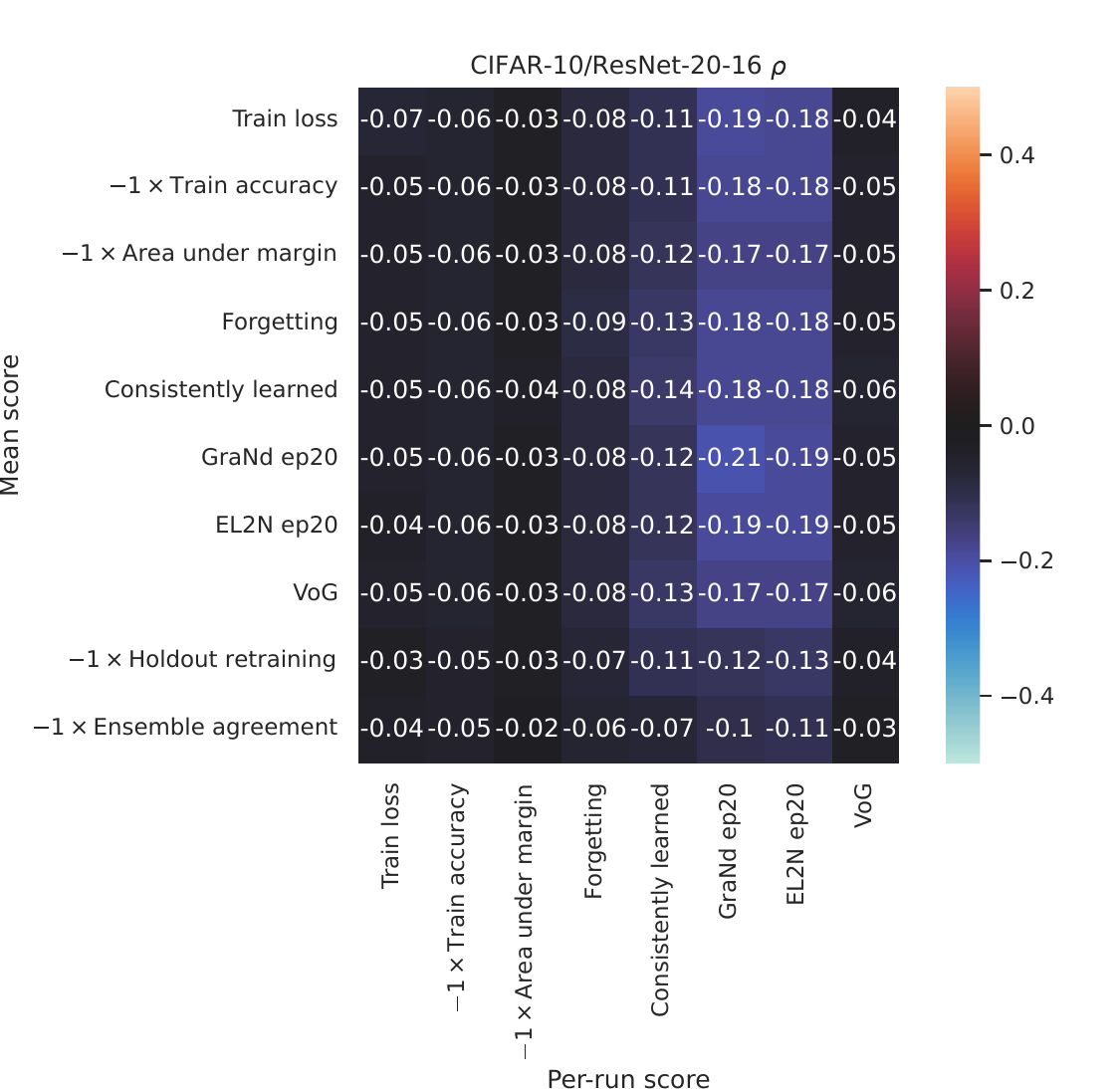}
    \includegraphics[width=0.32\textwidth]{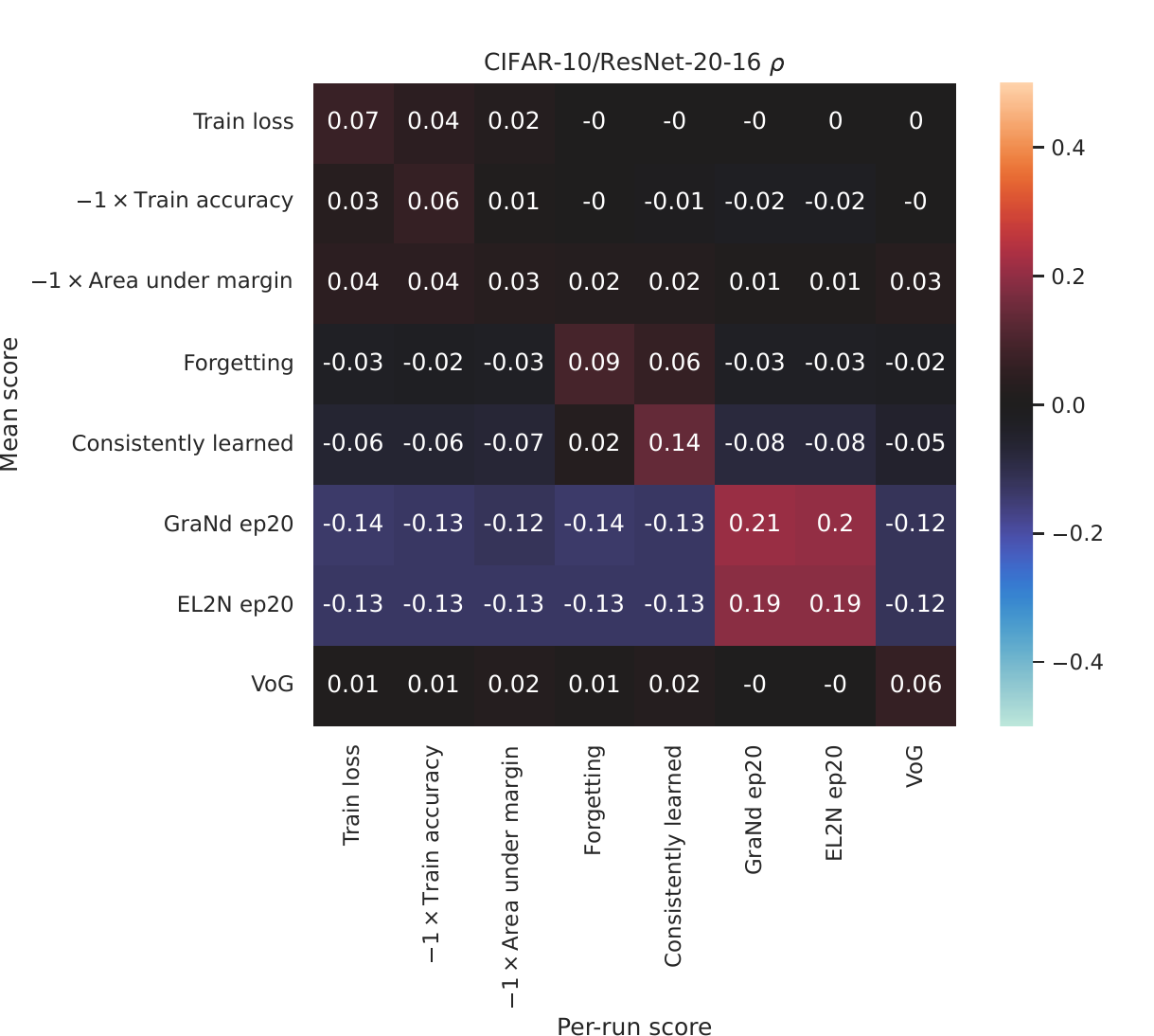}
    \caption{Spearman (rank) correlation between difficulty scores.
    To ensure positive correlations, scores are multiplied by $-1$ where required.
    \textbf{Left:} correlations between mean difficulty scores.
    All scores other than ``Ensemble agreement'' and ``Holdout retraining'' are averaged over 40 runs; the latter are pre-computed in \citet{carlini2019distribution},
    \textbf{Center:} difference in correlation when correlating mean scores (rows) with individual runs (columns), i.e. $\mathbb{E}_{Y}[Corr(\mathbb{E}_{X}[X], Y)] - Corr(\mathbb{E}_{X}[X], \mathbb{E}_Y[Y])$. Note that individual runs are not available for ``Ensemble agreement'' and ``Holdout retraining'' scores.
    \textbf{Right:} difference in correlation when correlating scores within individual runs scores, i.e. $\mathbb{E}_{X,Y}[Corr(X, Y)] - Corr(\mathbb{E}_{X}[X], Y)$.
    }
    \label{fig:covariance}
\end{figure}

\cref{fig:covariance} shows the correlation between mean scores, which is high for most scores with the lowest at $\rho > 0.6$ for \textsc{Ensemble agreement} \citep{carlini2019distribution}. 
\cref{fig:covariance} (Center) also shows the reduction/increase in correlation when using means to predict individual runs instead of other mean.
We find that all of these changes are negative, indicating that mean scores are less predictive of individual runs than other means, which is unsurprising given the high per-run variances observed in \cref{fig:variance}.
Note that the diagonal in \cref{fig:covariance} (Center) is $\rho - 1$, where $\rho$ is the correlation between the mean and individual runs of a given score.

Finally, \cref{fig:covariance} (Right) shows the reduction/increase in correlation when using individual runs to predict other runs instead of mean scores.
Note that the diagonal should be ignored since the score of an individual run is always perfectly correlated with itself---the diagonal in \cref{fig:covariance} (Right) is simply the inverse of \cref{fig:covariance} (Center), i.e. $1 - \rho$.
Positive values indicate that scores from an individual run are better predictors of another score than their averaged counterparts.
Although  most changes are negative, closely related scores such as \textsc{Train loss} and \textsc{Train accuracy}, \textsc{Forgetting} and \textsc{Consistently learned}, or \textsc{GraNd} and \textsc{EL2N} have a notable increase in correlation when considering individual runs.
This indicates that these pairs of scores vary in similar ways from run to run, so that they are more predictive of one another in a given run than than their averaged counterparts.

\begin{figure}
    \centering
    \includegraphics[width=0.48\textwidth]{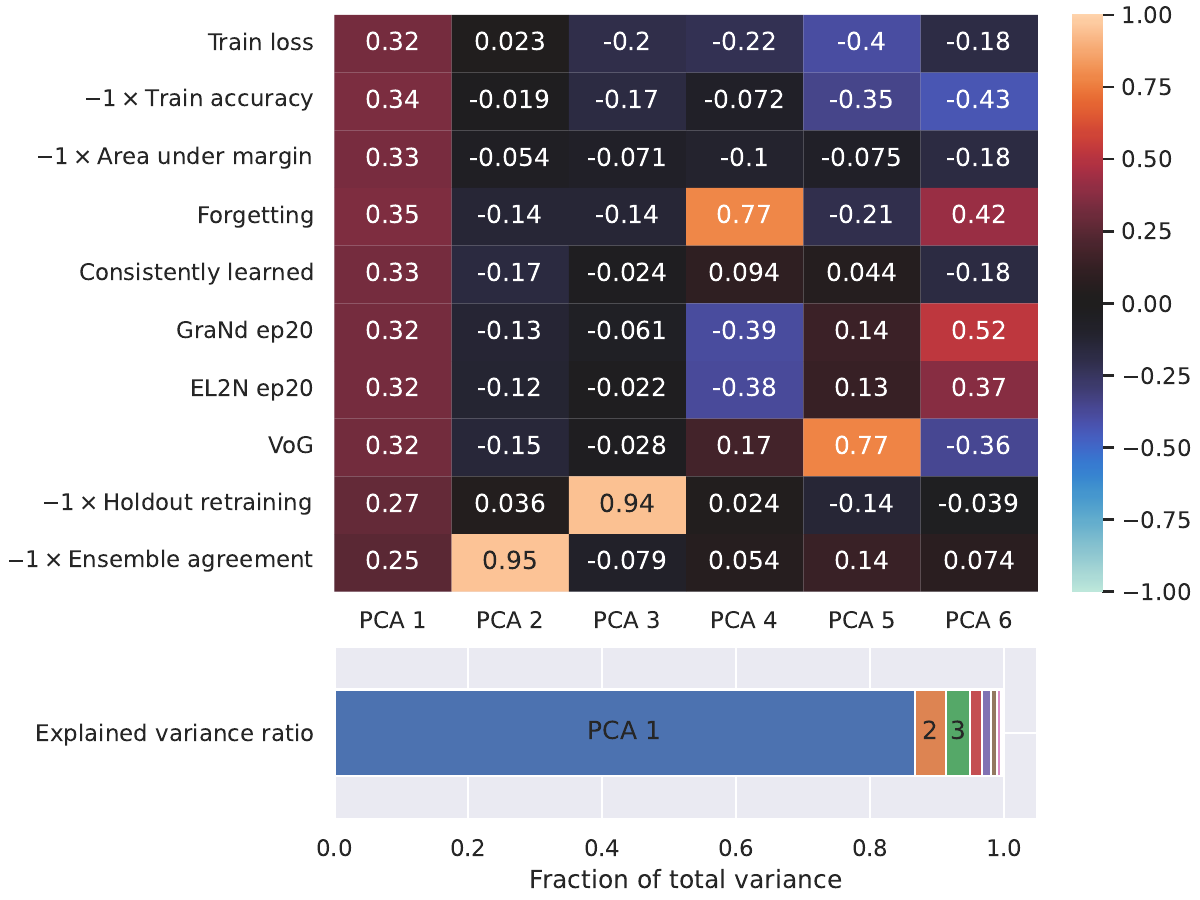}
    \includegraphics[width=0.48\textwidth]{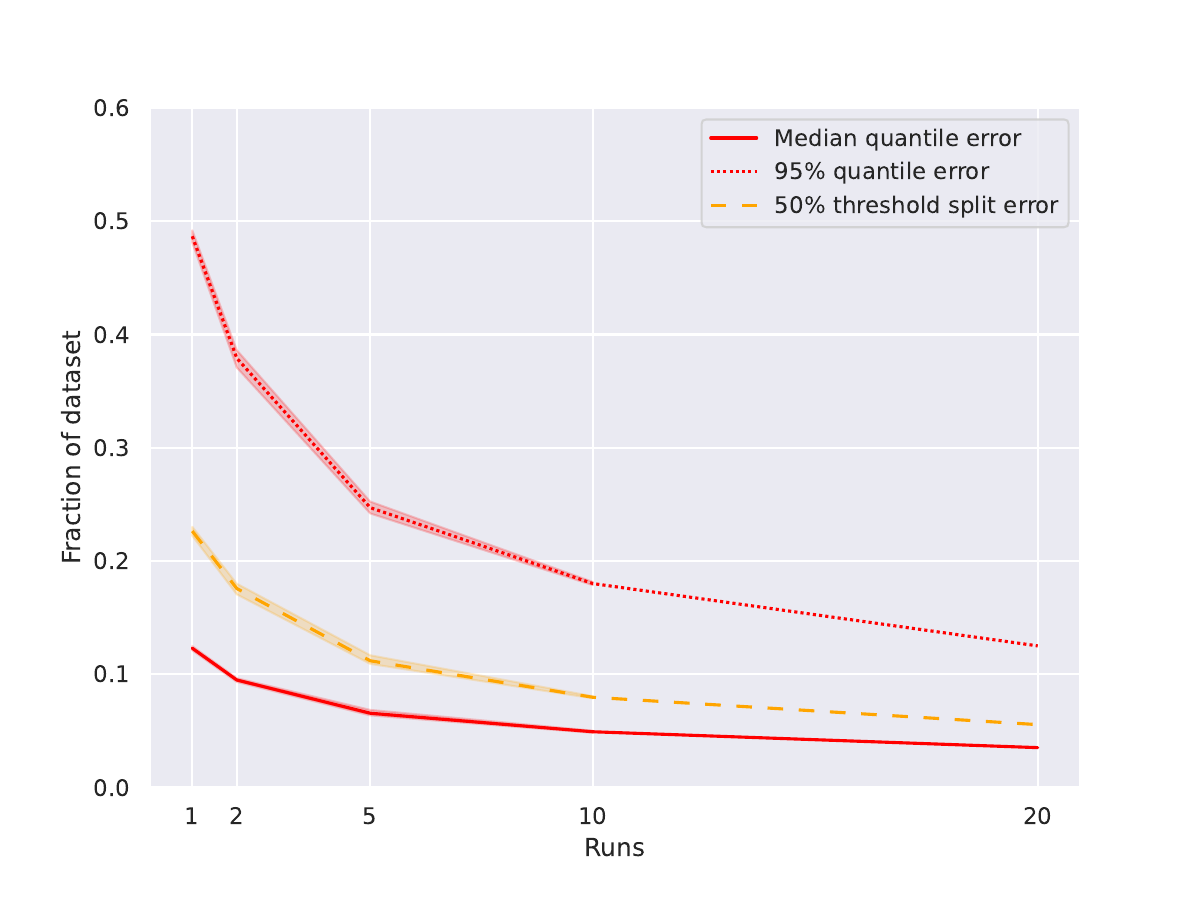}
    \caption{Principal component analysis of mean difficulty scores. All scores are first transformed to the normal distribution via quantile transform.
    \textbf{Left:} linear coefficients for top-6 principal components accounting for $>98\%$ of variation (heatmap top), and ratio of explained variance per component (stacked bars bottom).
    \textbf{Right:} variation in data splits by 50\% threshold, median absolute rank error, and 95\% confidence interval for ranks when using the top principal component as a score. This plot is analogous to \cref{fig:rank-threshold}.
    }
    \label{fig:pca-components}
\end{figure}

To determine the dimensionality of the space which scores span, we conduct a principal component analysis (PCA) in \cref{fig:pca-components} to extract the top directions of variation and the amount of variance present in each direction.
In the vein of the Spearman correlation used in \cref{fig:covariance}, we first apply a quantile transform to account for non-linear relationships among scores.
We find that the first principal component (PCA 1) accounts for the majority of variation (about 85\%), and is almost equally weighted across all scores.
This suggests that PCA 1 represents a common notion of an example's ``difficulty'' shared by all scores.

The remaining directions of variation seem to represent variation from specific scores, with \textsc{Ensemble agreement}, \textsc{Holdout retraining}, \textsc{Forgetting}, and \textsc{VoG} representing distinct directions that collectively capture about 10\% of total variation.
Note that this does not contradict findings in \citet{hooker2019compressed,baldock2021deep,carlini2019distribution} that suggest difficulty is 2 dimensional, since these works also consider different train/test splits (which we omit).

We also consider whether a linear combination of scores can reduce per-run variation.
\cref{fig:pca-components} (Right) plots the same ranking and thresholding analysis as \cref{fig:rank-threshold}, but using PCA 1 as a score.
We find a similar magnitude of variation as the average among the scores from \cref{fig:rank-threshold}, which reinforces the observation that the variation outside of PCA 1 is independent between different scores.

\subsection{Effect of Inductive Bias}
\label{sec:bias}

We examine the degree to which different model architectures, widths, and depths affect scores, and the degree to which such effects are distinguishable from the variation between runs.
Inspired by computational genetics and the genome-wide association study in particular \citep{clarke2011basic}, we run an independent two-sample test of difference of means for every example.
This effectively ranks examples by their bias to variance ratio in per-run scores, where bias is the change in difficulty between two different models.

\begin{figure}
    \centering
    \includegraphics[width=0.31\textwidth]{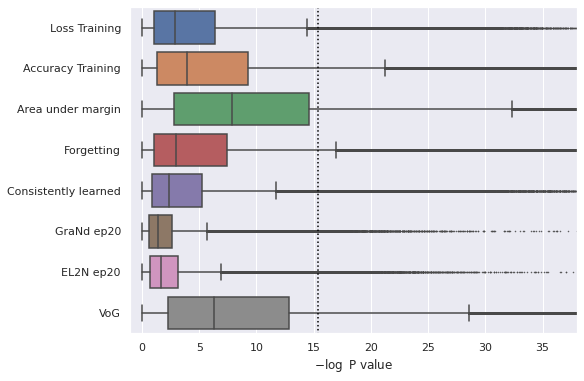}
    \includegraphics[width=0.34\textwidth]{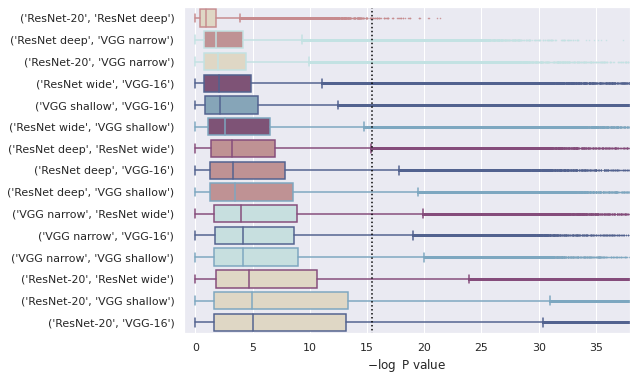}
    \includegraphics[width=0.31\textwidth]{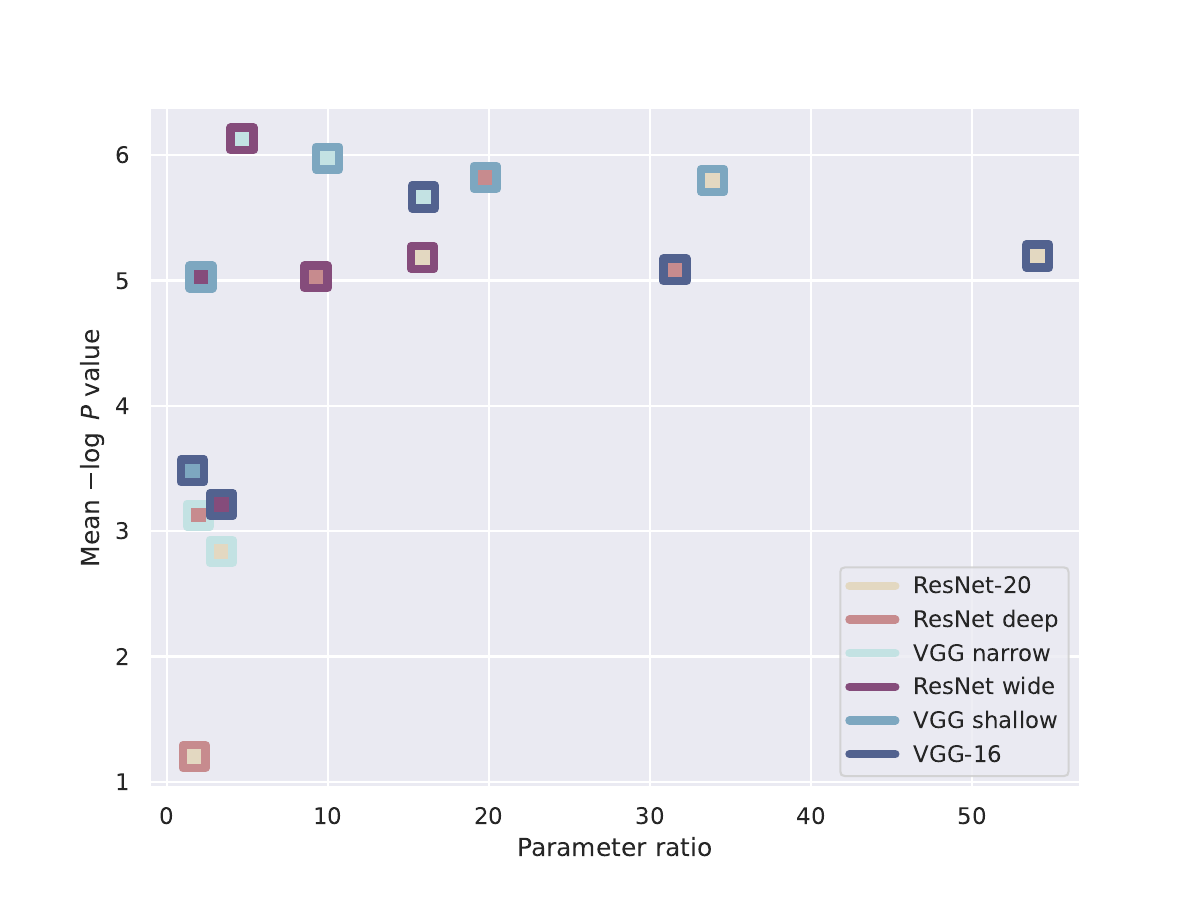}
    \caption{Distribution of $P$ values for unpaired t-tests for all scores and examples. $P$ values indicate the likelihood of the null hypothesis: namely that two populations of per-example difficulty scores from different inductive biases have the same mean. The negative log of the $P$ values is reported, with higher values corresponding to more statistically significant differences in means. The inductive biases compared are all pairs of models with different width, depth, and/or architecture (VGG/ResNet).
    \textbf{Left and Center:} $P$ value distribution by difficulty score or model architecture pairs. X-axis reports negative logarithm (higher values are more significant), with dotted line indicating a Bonferroni-corrected threshold of $p = 0.01$.
    \textbf{Right:} Geometric mean of $P$ value over all examples and scores, versus relative difference in model capacity (ratio of number of trainable parameters).}
    \label{fig:bias-pvalue}
\end{figure}

\cref{fig:bias-pvalue} plots the statistical significance of differences in mean difficulty for a variety of different pairs of model architectures.
We report the mean of the negative logarithm of $P$ values for all scores and/or all architecture pairs (equivalent to the geometric mean of $P$ values), with higher values corresponding to greater statistical significance.
Unsurprisingly, in \cref{fig:bias-pvalue} (Left) we find that scores that vary the least between runs \cref{fig:variance} tend to have higher statistical significance, such as \textsc{Area under margin}.
Interestingly, although \textsc{Loss training} and \textsc{Accuracy training} are closely related to \textsc{Area under margin}, they have much lower significance likely because they lose more of the information from the model output.
Among different architecture pairs in \cref{fig:bias-pvalue} (Center), we find large architectural changes (e.g. ResNet versus VGG) lead to more significant differences than small changes (e.g. increasing ResNet depth).
To investigate whether differences in model size (i.e. number of trainable parameters) can account for changes in example difficulty, we plot average $P$ value over all examples versus the size ratio between architecture pairs (\cref{fig:bias-pvalue} Right).
Although differences in model size correspond weakly to more significant differences in difficulty, there are also many pairs of similarly sized models which induce relatively significant changes in difficulty.

Finally, we show that a handful of the most significant examples can be used to fingerprint a model's inductive bias. \cref{fig:inductive-bias-examples} (Left) visualizes the examples with the most and least significant $P$ values (according to the geometric mean over all scores and model architecture pairs).
The top examples most sensitive to inductive biases are visibly more difficult than the bottom examples with consistent difficulty.
Scores from these examples are then used as features for identifying a model as either VGG-16 or ResNet-20, by fitting a logistic classifier (no regularization) to each type of score.
Classifers are fit using scores from 20 runs, and evaluated on a separate set of 20 runs.
\cref{fig:inductive-bias-examples} (Right) shows that for most scores, the top 8 examples are sufficient to perfectly classify VGG-16/ResNet model architectures, reliably beating 8 random examples, whereas the bottom 8 examples perform no better than chance.

\section{Discussion}

Despite appearing to be intuitive, the difficulty of an example is a rather nebulous concept that has thus far evaded systematic study.
In this work, we establish that there is a common notion of  difficulty among training examples, and that each example's difficulty is affected to different degrees by randomness (from model initialization, SGD training noise, etc.) and the method/model used to score it. 
We find that scores have high variance over individual runs which is inconsistent from example to example, raising concerns over the error inherent in score estimates.

The high correlation between scores suggests a greater degree of interchangeability among scores than one might expect.
This gives practitioners greater freedom in selecting scores, such as to replace a costly score like \citet{feldman2020neural} with a proxy as in \citet{jiang2020characterizing}, without requiring additional techniques as in \citet{liu2021understanding}.
This also tempers the advantages claimed by some scores as especially good predictors of other scores \citep{baldock2021deep, agarwal2020estimating}.

As the majority (85\%) of score variation among training examples lies in a single direction, we propose that this direction represents a common notion of ``difficulty'' measured by all scores.
Based on this finding, we hypothesize that the additional directions of variation proposed by \citet{hooker2019compressed, baldock2021deep, carlini2019distribution} also represent a second shared direction of variation measuring difficulty when an example is in the test set.
It is unclear if the second dimension in \citet{swayamdipta2020dataset} (based on variance over training) is related to these two directions.

Lastly, we show that examples range from being model-independent to being strongly dependent on the inductive biases of a given model.
The former examples allow scores to be compared between different models, whereas the latter can be used as lightweight features for classifying model architectures using a downstream method such as logistic regression.
In our case, a simple statistical test for difference of means was sufficient to identify model-dependent examples.
We note however, that there are methods from statistical genetics with greater statistical power (e.g. MTAG from \citet{turley2018multi}), which could potentially find examples that are sensitive to much smaller differences in inductive biases.
We leave the investigation of these methods for future work.

\newpage

\bibliographystyle{apalike}
\bibliography{citations}

This research was enabled in part by compute resources provided by Mila (mila.quebec).







\end{document}